\def\year{2022}\relax
\documentclass[letterpaper]{article} 
\usepackage{aaai23}  
\usepackage{times}  
\usepackage{helvet}  
\usepackage{courier}  
\usepackage[hyphens]{url}  
\usepackage{graphicx} 
\usepackage{natbib}  
\usepackage{caption} 
\DeclareCaptionStyle{ruled}{labelfont=normalfont,labelsep=colon,strut=off} 
\usepackage{amssymb}
\usepackage{amsmath}
\usepackage{color}
\usepackage{comment}
\usepackage{graphicx}
\usepackage{float}
\usepackage{subfigure}
\usepackage{algorithm}
\usepackage{algorithmic}
\usepackage{newfloat}
\usepackage{listings}
\floatstyle{ruled}
\newfloat{listing}{tb}{lst}{}
\floatname{listing}{Listing}

\title{Reinforcement Causal Structure Learning on Order Graph}
\author{
  Dezhi Yang\textsuperscript{\rm 1,2},
    Guoxian Yu\textsuperscript{\rm 1,2},
    Jun Wang\textsuperscript{\rm 2,*}
    Zhengtian Wu\textsuperscript{\rm 3},
    Maozu Guo\textsuperscript{\rm 4}
}
\affiliations{
    \textsuperscript{\rm 1}School of Software, Shandong University, Jinan, China\\
    \textsuperscript{\rm 2}SDU-NTU Joint Centre for AI Research, Shandong University, Jinan, China\\
    \textsuperscript{\rm 3}School of Electronic and Information Engineering, Suzhou University of Science and Technology, Suzhou, China\\
    \textsuperscript{\rm 4}College of Elec. \& Inf. Eng., Beijing University of Civil Engineering and Architecture, Beijing, China\\
    dzyang@mail.sdu.edu.cn, \{gxyu, kingjun\}@sdu.edu.cn, wzht8@mail.usts.edu.cn, guomaozu@bucea.edu.cn
}

\begin{document}

\maketitle

\begin{abstract}
Learning directed acyclic graph (DAG) that  describes the causality of observed data is a very challenging but important task. Due to the limited quantity and quality of observed data, and non-identifiability of causal graph, it is almost impossible to infer a single precise DAG. Some methods approximate the posterior distribution of DAGs to explore the DAG space via Markov chain Monte Carlo (MCMC), but the DAG space is over the nature of super-exponential growth, accurately characterizing the whole distribution over DAGs is very intractable. In this paper, we propose {Reinforcement Causal Structure Learning on Order Graph} (RCL-OG) that uses order graph instead of MCMC to model different DAG topological orderings and to reduce the problem size. RCL-OG first defines reinforcement learning with a new reward mechanism to approximate the posterior distribution of orderings in an efficacy way, and uses deep Q-learning to update and transfer rewards between nodes. Next, it obtains the probability transition model of nodes on order graph, and computes the posterior probability of different orderings. In this way, we can sample on this model to obtain the ordering with high probability. Experiments on synthetic and benchmark datasets show that RCL-OG provides accurate posterior probability approximation and achieves better results than competitive causal discovery algorithms.
\end{abstract}

\section{Introduction}
Causal discovery can reveal the causal mechanisms underlying natural phenomena, it has been attracting the attention of many disciplines \cite{pearl2009causality}. The golden standard for causal discovery is to conduct randomized intervention experiments, which is however restricted by various factors (i.e., cost and ethics) and even infeasible. Therefore, inferring causal structure from passively observable data is more attractive and the only road for causal discovery \cite{pearl2009causality,peters2017elements}.

The causal graph captures the underlying causal mechanism of the data, where nodes and edges correspond to variables and causality between variables, respectively. The causal graph typically is a directed acyclic graph (DAG). To find causal graphs, score-based methods assign a score $S(\mathcal{G})$ to each DAG $\mathcal{G}$ and search the DAG with the best score in the DAG space:
\begin{equation}
    \min_{\mathcal{G}} S(\mathcal{G}),\ \text{s.t.} \ \mathcal{G}\in DAGs
    \label{eq1}
\end{equation}%
However, due to the combinatorial nature of acyclic constraints, the search complexity  is generally considered to be NP-hard \cite{chickering2004large}. Most methods employ heuristic strategies. To name a few, GES  \cite{chickering2002optimal} greedily adds edges that increase the score of DAG the most; and max-min hill climbing \cite{tsamardinos2006max} searches DAGs in the space reduced by constraint-based methods. Some other methods align learning an optimal DAG with learning an optimal DAG topological ordering, and search causal structures in a smaller ordering space \cite{teyssier2005ordering,raskutti2018learning,bernstein2020ordering}.

Most score-based methods can only provide a single DAG or its Markov equivalent class with the best score, which may produce poor results, especially in the case of limited data, poor data quality, and unidentifiable causal graphs. Different from score-based methods \cite{tsamardinos2006max,huang2018generalized}, MCMC-based methods \cite{madigan1995bayesian,niinimaki2016structure,deleu2022bayesian} infer the posterior probability of the DAG structure $P(\mathcal{G}|\mathcal{D})$ from the observation dataset $\mathcal{D}$ to account for the epistemic uncertainty over graph structures. However, too huge DAG combination space makes it very difficult to characterize the DAG posterior distribution via MCMC and even produces slow mixing. 

\begin{figure}[ht]
\centering
\includegraphics[scale=0.75]{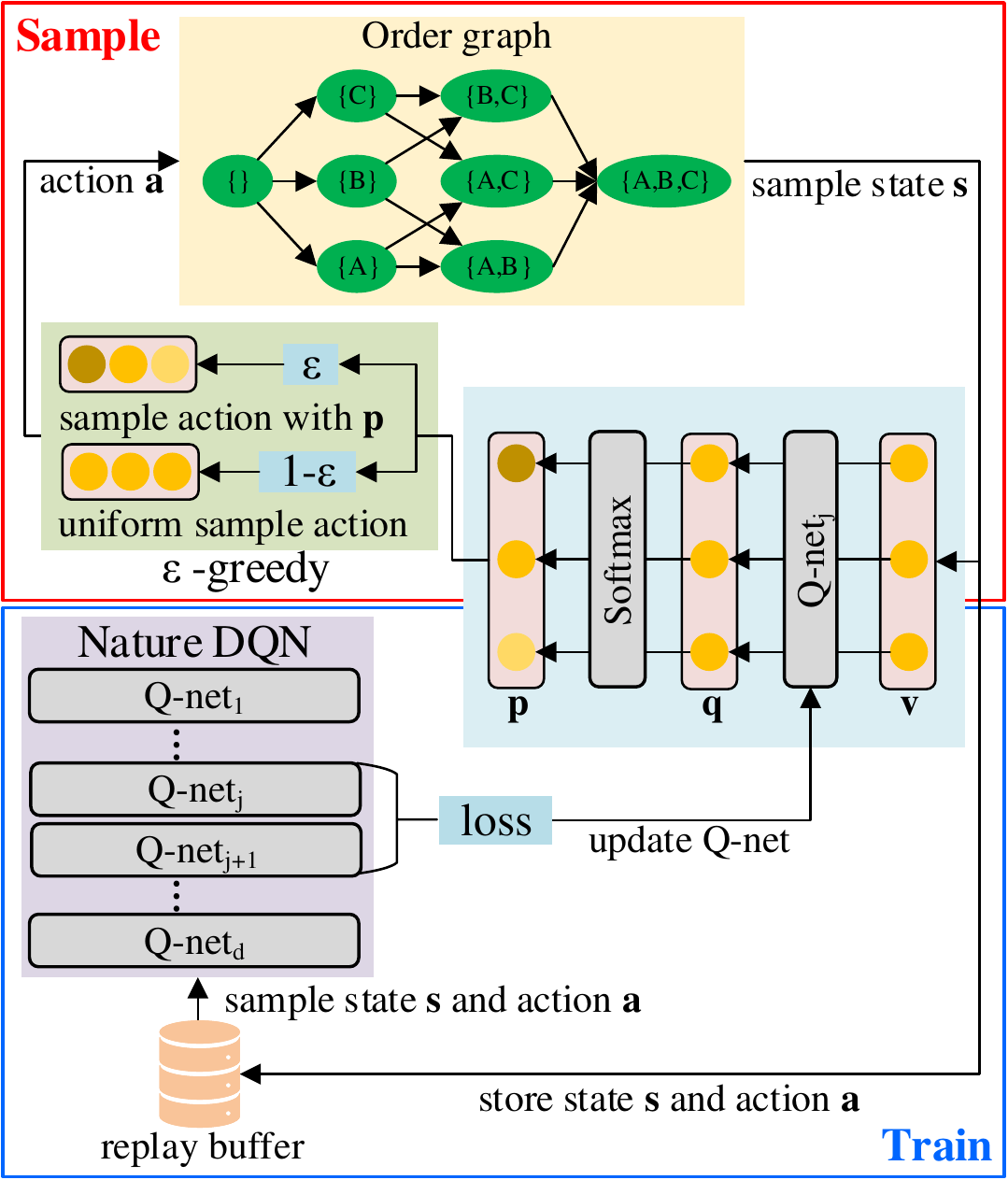} 
\caption{Schematic framework of RCL-OG. In the sampling phase, RCL-OG computes the transition probabilities $\mathbf{p}$ by Q-nets, uses the $\epsilon$-greedy policy to sample states and actions, and stores them in the replay buffer. During the training phase, it draws a batch of samples from the replay buffer to update Q-nets via Nature DQN strategy. The trained Q-nets will be used as the probability transition model to encode possible ordering space.}
\label{fig1}
\end{figure}

In this paper, we propose a deep reinforcement learning based solution (RCL-OG) that approximates the posterior probability of DAG topological orderings using order graph. The order graph can clearly capture the generation process of each kind of ordering, and sampling the order can be regarded as a Markov decision process, starting from the empty set, by selecting one variable to join the set based on the ordered variables. By designing new transition rewards for the nodes of the order graph, we design the formula for calculating the probability of transition between nodes and prove the detailed-balance condition. We show that computing the transfer rewards is a dynamic programming problem, so we use the {Nature DQN \cite{mnih2015human}} strategy with a new target function to approximate the transfer rewards, as outlined in Figure \ref{fig1}. 

RCL-OG samples nodes in the order graph as the state input to the Q-nets, selects actions according to the output of the Q-nets and $\epsilon$-greedy strategy, and then sends them to the order graph to get next states. RCL-OG stores the sampled states, actions and next states into the experience replay pool, and updates the Q-network by Nature DQN strategy. Next, it takes the trained Q-network as the desired probability transition model on the order graph. In this way, RCL-OG can learn the posterior distribution of DAG topological orderings in an efficacy way and discover causal structures by sampling on the trained probability transition model. We evaluate RCL-OG on synthetic and benchmark datasets. RCL-OG can find the correct causal structures with high probability, which are better than those found by traditional structure learning methods \cite{spirtes2000causation,kuipers2017partition,zhu2019causal}.

The main contributions of our work are listed as follows:\\
{
(i) 
We introduce an order graph with a much smaller search space to improve the efficacy of causal structure search, and define states, actions and rewards for reinforcement learning based search on this graph.\\
(ii) 
We introduce neural networks to approximate the transition probability on the order graph, and reformulates  orderings search and  probability model training into the reinforcement learning framework to leverage their advantages for learning the posterior probability distribution of DAG topological ordering.\\
(iii) Experiments on simulated datasets with known transition probability confirm the reliability of RCL-OG on inferring the transition probability. By comparing with state-of-the-art methods on simulated and real-word datasets, we prove that RCL-OG can sample on learned probability model to accurately find  causal structures.}

\section{Related Work}
Traditional causal structure learning methods can be divided into two categories: constraint-based and score-based methods. Constraint-based methods obtain the skeleton of causal graph by testing the conditional independence between variables, and then orient the skeleton to identify Markov equivalence classes. {PC \cite{spirtes2000causation} is the earliest and most classical constraint-based method.} \citet{sun2007kernel} and \citet{zhang2011kernel} further applied kernel-based conditional independence test to improve the accuracy. Constraint-based methods require sufficient samples to ensure the reliability of the conditional independence tests, but the observed samples for causal structure learning are often insufficient, which limit the application of constraint-based methods.

Score-based methods search for DAGs using various well-defined score functions (i.e. Bayesian Information Criterion (BIC) \cite{schwarz1978estimating} and Minimum Description Length (MDL)  \cite{,chickering2002optimal}). But the combinatorial nature of search makes it difficult for these methods to explore large graph structures. Some methods attempt the continuous optimization to learn causal structure from data. \citet{zheng2018dags} formulated the combinatorial optimization problem as a continuous optimization problem by introducing a smooth characterization for the acyclicity. Gran-DAG \cite{lachapelle2019gradient} and NOTEARS-MLP \cite{zheng2020learning} extend the smooth acyclic constraints to neural networks for nonlinear data. Some other methods (i.e. CGNN \cite{goudet2018learning} and SAM \cite{kalainathan2018structural}), also use neural networks for causal discovery, but they do not guarantee acyclicity.

Most causal structure learning methods return a single DAG, which are limited by observable data and often fail to produce good results. Markov chain Monte Carlo (MCMC)-based methods approximate a whole posterior distribution over DAGs by exploring the DAG space. In essence, MCMC is a sampler that simulates a Markov chain in the DAG space by local transitions (e.g. adding or removing edges) \cite{madigan1995bayesian}. \citet{friedman2003bayesian} proposed an MCMC sampler in the space of node orders to reduce the search space and avoid slow mixing, but also introduced a bias. \citet{kuipers2017partition} further refined these methods by modifying the underlying space or local transition of Markov chains. DAG-GFlowNet \cite{deleu2022bayesian} used generative flow network instead of MCMC and approximated posterior of DAGs with deep learning. {A smaller and well-defined search space can help MCMC-based methods better sample possible structure. We propose to replace MCMC with order graphs to reduce the underlying search space, and model the transition between orderings as a Markov decision process to improve the search performance of RCL-OG.}

RL has also made good progress in causal structure learning \cite{khalil2017learning}. \citet {bello2016neural} used the simple reward mechanisms of RL to solve combinatorial optimization problems. RL-BIC \cite{zhu2019causal} uses deep-RL \cite{sutton2018reinforcement} to search for a single high score structure, but it neglects the decision process. {Exploring the search space is essential for causal structure learning, but we typically can only explore a small part of it, due to the high search complexity and huge space. In this work, we use neural networks to approximate the posterior probability of DAG topological ordering, and reformulate the exploration and learning into the RL framework to effectively leverage the searched structure knowledge and infer the posterior distribution of the search space.
}

\section{Background}

\subsection{Structure Learning}

Given a causal graph $\mathcal{G}$ with $d$ random variables $\mathcal{X}=\{X_1,\cdots,X_d\}$, whose joint distribution can be decomposed according to $\mathcal{G}$:
\begin{equation}
    P(X_1,\cdots,X_d)=\prod_{k=1}^d P(X_k|\mathrm{Pa}(X_k))
    \label{eq2}
\end{equation}%
where $\mathrm{Pa}(X_k)$ is the set of parents of $X_k$ in $\mathcal{G}$. We assume that the observed dataset $\mathcal{D}$ is generated by Structural Equation Model (SEM) with additive noises: $x_i=f_i(Pa(x_i))+\epsilon_i, i=1,2,\cdots,d$. We also assume causal minimality, namely each function $f_i$ is not a constant \cite{peters2014causal}. 

The problem of finding a DAG can be cast as finding a variable ordering \cite{teyssier2005ordering}. We can easily represent the ordering as a fully connected DAG by setting frontal nodes in the order as parents of subsequent nodes. Each DAG can be consistent with more than one orderings, so searching the real causal graph $\mathcal{G}^*$ can be regarded as searching an ordering, whose fully connected DAG is a super-graph of $\mathcal{G}^*$.

Given a dataset $\mathcal{D}$, we can infer the posterior distribution of DAGs over data. A natural idea is that the probability of a DAG is proportional to its score. If the score function can be decomposed, we have the following:
\begin{equation}
    \displaystyle
    P(\mathcal{G}|\mathcal{D})\propto R(\mathcal{G})=\prod_{k=1}^d S(X_k,\mathrm{Pa}(X_k)|\mathcal{D})
    \label{eq3}
\end{equation}%
where $S(\cdot)$ is the decomposed partial score function. We can also apply the ordering to the above formula: $P(\mathcal{L}|\mathcal{D})\propto\prod_{k=1}^d S(X_k,\mathrm{U}(X_k)|\mathcal{D})$, where $\mathcal{L}$ stands for an ordering and $\mathrm{U}(X_k)$ is the set of variables ahead $X_k$ in $\mathcal{L}$.

\subsection{Order Graph}
The order graph is introduced to describe the generation process of variable orderings, and its structure is exampled in the top of Figure \ref{fig1}, where nodes represent the set of sorted variables and edges means adding a variable to the set of its head nodes. We define nodes on the order graph as states $s\in\mathcal{S}$ and edges as actions $a\in\mathcal{A}$. Different paths from the top empty set node $s_0$ to the bottom universal set node $s_d$ capture different variable orderings. Therefore, determining a path is a Markov decision process.

\section{The Proposed Methodology}

In this section, we first introduce how to model the whole variables ordering space with the order graph. We design new definitions of states, rewards and actions for the order graph, provide the formula of transition probability between states and prove the detailed-balance condition. Then we discuss how to use RL to estimate the transition probability and obtain the posterior distribution of DAG topological orderings.

\subsection{Modeling Variable Orderings}
\textbf{States and rewards}:
The order graph is a DAG, in which there are only forward transitions (add variables to sorted set) and no backward transitions. To enable the order graph to model transitions between complete orderings, we modify its node $s$ meaning to be the set of orderings corresponding to paths through $s$ (not all paths, as we will explain later). Take the order graph in Figure \ref{fig1} for example, node $s=\{A\}$ denotes the set of orderings $O(s)=\{[A,B,C],[A,C,B]\}$. Consider the transition from state $s$ to its sub-state $s'$, since $s'$ is generated by $s$, the set of orderings represented by $s'$ only includes the orderings corresponding to paths that pass through both $s$ and $s'$, namely $O(s')\subset O(s)$. In addition, we assume that there exists a definite path $l$ from state $s_0$ to $s$, so that the size of ordering sets represented by $s$ and $s'$ is further reduced, because the path corresponding to the orderings in them must go through $l$.  Under these assumptions, we define the reward of state $s$ as the sum of the scores of all orderings it represents: $R(s)=\sum_{\mathcal{L}\in O(s)}R(\mathcal{L})$, where $R(\mathcal{L})$ is the score of ordering $\mathcal{L}$.

\begin{figure*}
\centering
\includegraphics[scale=1.04]{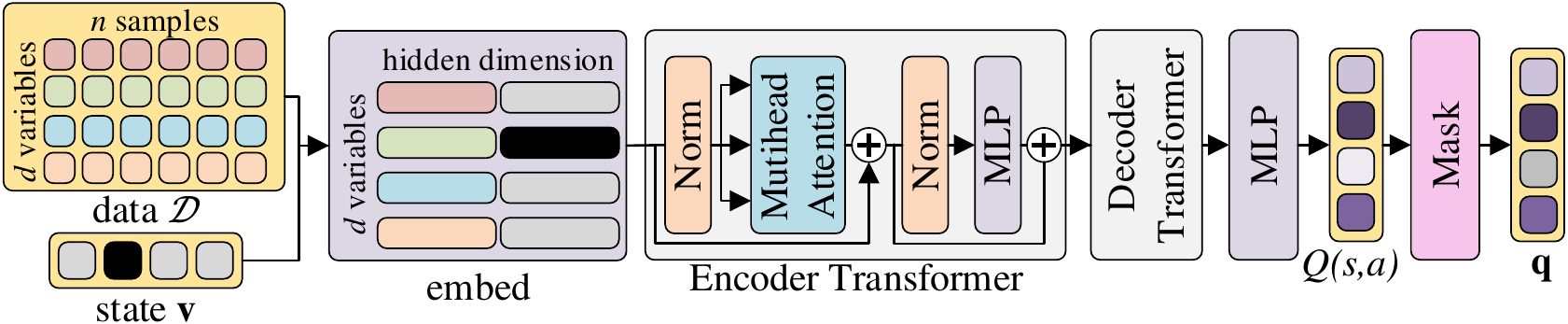} 
\caption{The architecture of Q-net. The input data is a data matrix of $n$ samples represented by $d$ variables. The input state is a binary vector of length $d$. Each variable is embedded through the concatenation of the embeddings of its corresponding sample values and the embeddings of its corresponding entry in the state vector. These embeddings are fed into a Transformer and encoded as features. The decoder has the same structure as the encoder, along with a multi-layer perception to decode the features into $Q(s,a)$ of the state $s$ for each possible action $a$. Finally, $Q(s,a)$ is masked as $\mathbf{q}$ to avoid explored actions.}
\label{fig2}
\end{figure*}

\textbf{Detailed-balance condition}:
{After encoding states as ordering sets, the probability of state $s$ is given by:
\begin{equation}
    \displaystyle
    P(s)={\sum}_{\mathcal{L}\in O(s)}P(\mathcal{L})
    \label{eq4}
\end{equation}%
where $O(s)$ is the set of orderings encoded by $s$ and $P(\mathcal{L})$ is the probability of ordering $\mathcal{L}$.} Under the assumption that there exists a definite path from state $s_0$ to $s$, since $O(s')\subset O(s)$, we get the following transition probability:
\begin{equation}
    \displaystyle
    P(s'|s)=\frac{P(s's)}{P(s)}=\frac{P(s')}{P(s)}=\frac{\sum_{\mathcal{L'}\in O(s')}P(\mathcal{L'})}{\sum_{\mathcal{L}\in O(s)}P(\mathcal{L})}
    \label{eq5}
\end{equation}%
Notice that $P(\mathcal{L}|\mathcal{D})\propto R(\mathcal{L})$, we have $P(s'|s)=R(s')/R(s)$ and find the transition of states on the order graph satisfying the detailed-balance condition:
\begin{equation}
    \displaystyle
    P(s'|s)R(s)=P(s|s')R(s')
    \label{eq6}
\end{equation}%
where $P(s|s')=1$, because $s'$ is produced by $s$.

\textbf{Actions and values}:
{The calculation of state rewards $R(s)$ may be difficult because of the high complexity of ordering rewards $R(\mathcal{L})$. As such, we need to simplify the formula of transition probability.} In the order graph, there is a unique action between each pair of parent and child states, which means putting a variable into the sorted set to make state $s$ transiting to $s'$. We define the reward of action $a$ as decomposed partial score function:
\begin{equation}
    \displaystyle
    R(a)=S(s'-s,s|\mathcal{D})
    \label{eq7}
\end{equation}%
where $s'-s$ is the difference between two sets with sorted variables. We can express the ordering reward as the product of action rewards and then compute the sum of ordering rewards to obtain the rewards for states. We take the action reward into the detailed-balance condition and get the following formulas:
\begin{equation}
    \begin{aligned}
    \displaystyle
    &P(s'|s)=\frac{Q(s,a)}{\sum_{a'\in\mathcal{A}_s}Q(s,a')}\\
    Q(s,a)=&R(a)\sum_{a'\in\mathcal{A}_{s'}}Q(s',a'),\;Q(s_{d-1},a)=R(a)\\
    \end{aligned}
    \label{eq8}
\end{equation}%
where $a$ is the action between $s$ and $s'$, $\mathcal{A}_s$ is the set of all possible actions of $s$, and $s_{d-1}$ stands for states in the second-to-last layer of the order graph with only one action. The above formulas consider that the transition probability from state $s$ to $s'$ is independent of the paths (or actions) before $s$ and can be obtained by recursively calculating $Q(s,a)$. We call $Q(s,a)$ the value of doing action $a$ in state $s$. By computing $Q(s,a)$ for all states and their actions, we can obtain the transition probabilities between each pair of states.

\subsection{Learning Probability Transition Model}

\textbf{Network structure}:
Although the order graph greatly reduces the size of the state-action space compared to MCMC, it is still difficult to traverse all states and actions to compute the  transition probability between each pair of states. RL simplifies the search process based on Markov decision and guides us to explore important parts of the state-action space on the order graph based on rewards and defined target function (which usually prefers higher rewards for states and actions). In the RL framework, we use neural networks to parameterize $\mathrm{log}Q(s,a)$ (we will explain why $\mathrm{log}$ later) {and to borrow their capacity for generalizing to states and actions not explored by RL.} In practice, we adopt an encoder-decoder structure, as shown in Figure \ref{fig2}.

The state is expressed as a binary vector $\mathbf{v}\in \mathbb{R}^d$, whose entry is 1 if the corresponding variable is included in the set of sorted variables represented by the state, and 0 otherwise. We embed the data $\mathcal{D}$ with the state vector $\mathbf{v}$ and map the embedding into a feature vector through a Transformer \cite{vaswani2017attention}. Next, we use another transformer to decode the feature vector into a value vector $\mathbf{q}$:{
\begin{equation}
    \displaystyle
    \mathbf{q}=\mathrm{Transformer_{de}}(\mathrm{Transformer_{en}}(\mathrm{embed}(\mathcal{D}, \mathbf{v})))
    \label{eq9}
\end{equation}%
Each entry of $\mathbf{q}$ denotes $\mathrm{log}Q(s,a)$ of the corresponding action. But for a state $s$, not all actions are feasible. For example, it is not possible for $s=\{A\}$ to add variable $A$ to the set again, so we mask impossible actions of $\mathbf{q}$ as $-9e15$. We call the entire network as \emph{structure Q-net}.}

\textbf{Reinforcement learning based ordering search}:
{As shown in Figure \ref{fig1},} starting from the initial state $s_0$ of the order graph, we first input the binary vector $\mathbf{v}_0$ of $s_0$ into Q-net to obtain the value vector $\mathbf{q}$. We perform softmax to process $\mathbf{q}$ into a probability vector $\mathbf{p}$, and then use the $\epsilon$-greedy strategy to select the action. Specifically, we sample possible action by $\mathbf{p}$ with probability $\epsilon$, and uniformly sample action with probability $1-\epsilon$. We then perform the sampled action on the order graph to get the next state, and repeat the above process. When reaching the state $s_d$, we start the above sampling process from $s_0$ again until a certain number of iterations. We store the sampled state $s$, action $a$ and next state $s'$ into a replay buffer. At each iteration, we randomly draw  a batch of samples from the buffer and compute their current and target value to update Q-net.

According to Eq. \eqref{eq8}, we need to calculate action rewards to express $Q(s,a)$. We can easily compute the logarithm of the action reward by score function:
\begin{equation}
    \displaystyle
    \mathrm{log}R(a)=\mathrm{LocalScore}(s'-s,s|\mathcal{D})
    \label{eq10}
\end{equation}%
where the score function is Bayesian Information Criterion (BIC), which is not only {consistent but also decomposable} \cite{chickering2004large}. {Since the score function computes the logarithm of reward $R(a)$, we rewrite the target function of Q-net as the logarithm of $Q(s,a)$ for ease calculation:
\begin{equation}
    \displaystyle
    y(s,a)=
    \begin{cases}
    \mathrm{log}R(a)& \text{$s'=s_d$} \\
    \begin{aligned}
    \mathrm{log}R(a)&+\\
    &\mathrm{log}(\sum_{a'\in \mathcal{A}_{s'}}\mathrm{exp}(Q(s',a')))
    \end{aligned}
    & \text{$s'\neq s_d$}
    \end{cases}
    \label{eq11}
\end{equation}%
For ease understanding, we still use $Q(s,a)$ to denote the output of Q-net.} We use the mean square error between  current output value  and the calculated target value as the loss: $loss=\frac{1}{m}\sum^m_{j=1}(y(s,a)-Q(s,a))^2$, where $m$ is the size of batched samples drawn from the replay buffer.

Inspired by Nature DQN \cite{mnih2015human}, we use different Q-nets to parameterize $Q(s,a)$ in different layers of the order graph. For an order graph with $d$ variables, it has $d+1$ layers, indexed from $1$ to $d+1$. We use the same architecture to initialize $d$ Q-nets corresponding to layer $1$ to $d$ of the order graph, respectively. {As shown in Figure \ref{fig1}, the state-action values of state $s$ at $j$-th layer and its sub-state $s'$ at $(j+1)$-th layer are respectively parameterized by the $j$-th Q-net and the $(j+1)$-th Q-net to calculate $loss$}, which not only decouples the current value and target value, but also reduces the number of state-action values that Q-net need to parameterize.

\section{Results and Analysis}
\subsection{Comparison with exact transition probability}
Since the posterior probability is the product of transition probabilities, we compare the transition probability calculated by the networks against the exact ones. If the networks give the exact probability, then RCL-OG can provide accurate posteriors of DAG topological orderings. However, computing the exact transition probability between states on the order graph still requires to exhaustively enumerate all possible orderings and compute their scores, which is only feasible for graphs with a few variables. Therefore, we perform experiment on a DAG generated with 5 nodes and $2$$\times$$5$ edges under the Erd{\"o}s-R{\'e}nyi model \cite{erdos1960evolution}. We generate 200 samples based on linear model with Gaussian noise. By enumerating all possible orderings of these variables and computing their scores, we can obtain the exact transition probability. 

For state $s$ and its sub-state $s'$ on the order graph, we no longer assume that there exists a definite path from $s$ to $s'$. So the reward for $s$ is the sum of scores for all paths that pass through $s$, and this for $s'$ is the sum of scores for all paths that go through both $s$ and $s'$. The exact transition probability from $s$ to $s'$ is:
\begin{equation}
    \displaystyle
    P(s'|s)=\frac{\sum_{\mathcal{L'}\in O_{all}(s')}R(\mathcal{L'})}{\sum_{\mathcal{L}\in O_{all}(s)}R(\mathcal{L})}
    \label{eq12}
\end{equation}%
where $O_{all}(s')$/$O_{all}(s)$ {include all orderings that go through state $s'$/$s$.} We input the state vector of $s$ to Q-nets and execute softmax on the value vector outputted by networks to obtain the probability vector, then the value of $s'$ on the probability vector is the estimated transition probability from $s$ to $s'$.

\begin{figure}
\centering
\includegraphics[scale=0.21]{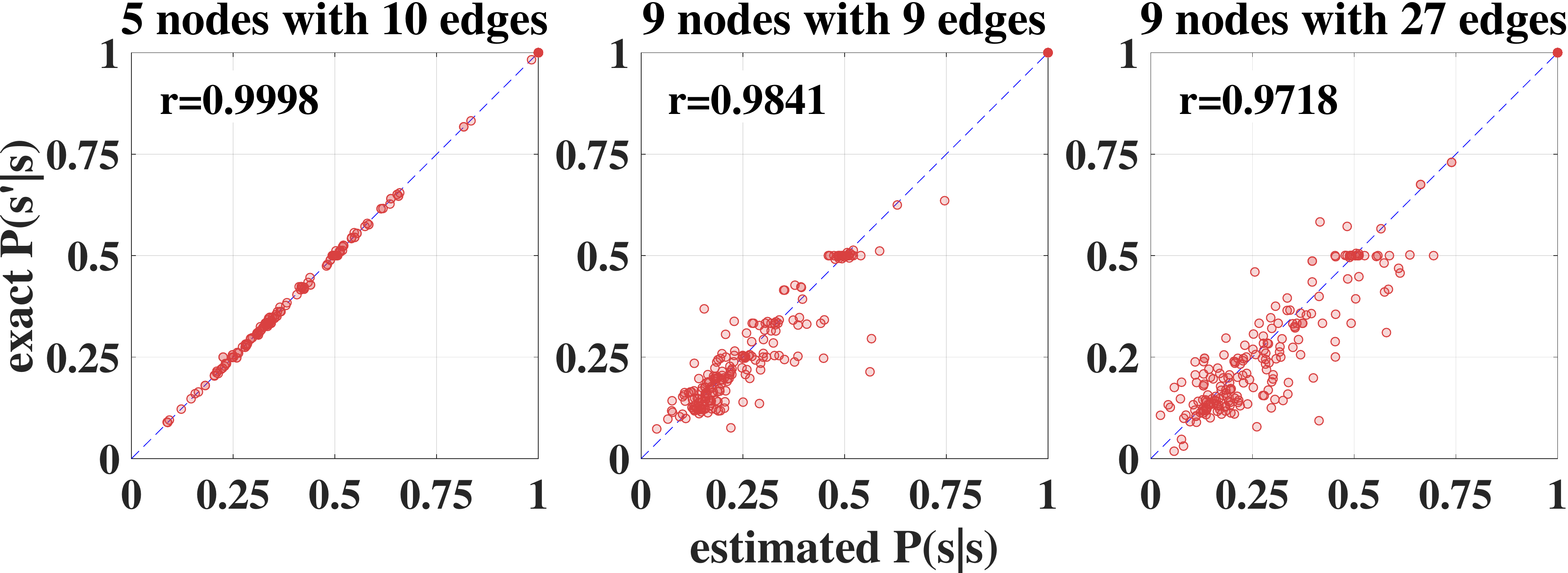} 
\caption{Distribution of exact transition probabilities (y-axis) and the ones estimated by RCL-OG (x-axis). We perform experiments with 5 different DAGs, randomly select 50 pairs of states and sub-states for each DAG. $r$ is the Pearson correlation coefficient.}
\label{fig3}
\end{figure}

RCL-OG can accurately learn the transition probability between states on order graph, and  give precise approximation of the posterior distribution over DAG topological orderings $P(\mathcal{L}|\mathcal{D})$. For validation, we randomly select 250 pairs of states and sub-states from the order graph of the experimental DAGs,  plot the exact probabilities and the ones estimated by RCL-OG in Figure \ref{fig3}. We see the exact and estimated probabilities are strong linearly correlated.

We also repeat the above experiments on the DAGs with $d$$=$$9$ nodes and $1d,3d$ edges, and reveal results in Figure \ref{fig3}. RCL-OG does not perfectly estimate all transition probabilities as $d=9$, which we believe is caused by the size of states and actions space. With the increase of variables, the space of states and actions in the order graph super-exponentially increases, and it is impossible to search all possible states and actions in the sampling phase. Although RCL-OG cannot accurately approximate the transition probability {that is infrequent or not searched}, experimental results at multi-nodes graph show that this deficiency does not prohibit it from discovering the exact causal structure. For a state $s$ in the order graph, the reward gap between different actions enlarges with the increase of variables, and the correct action tend to have a higher value than other actions, which also make the transition probability of the correct action much higher than other actions. In the sampling phase, the correct action is more likely to be sampled and used to train the networks, so RCL-OG is still able to find the exact causal structure in the multi-nodes graph.

\subsection{Experiments on simulated data}
\begin{figure}
\centering
\includegraphics[scale=0.42]{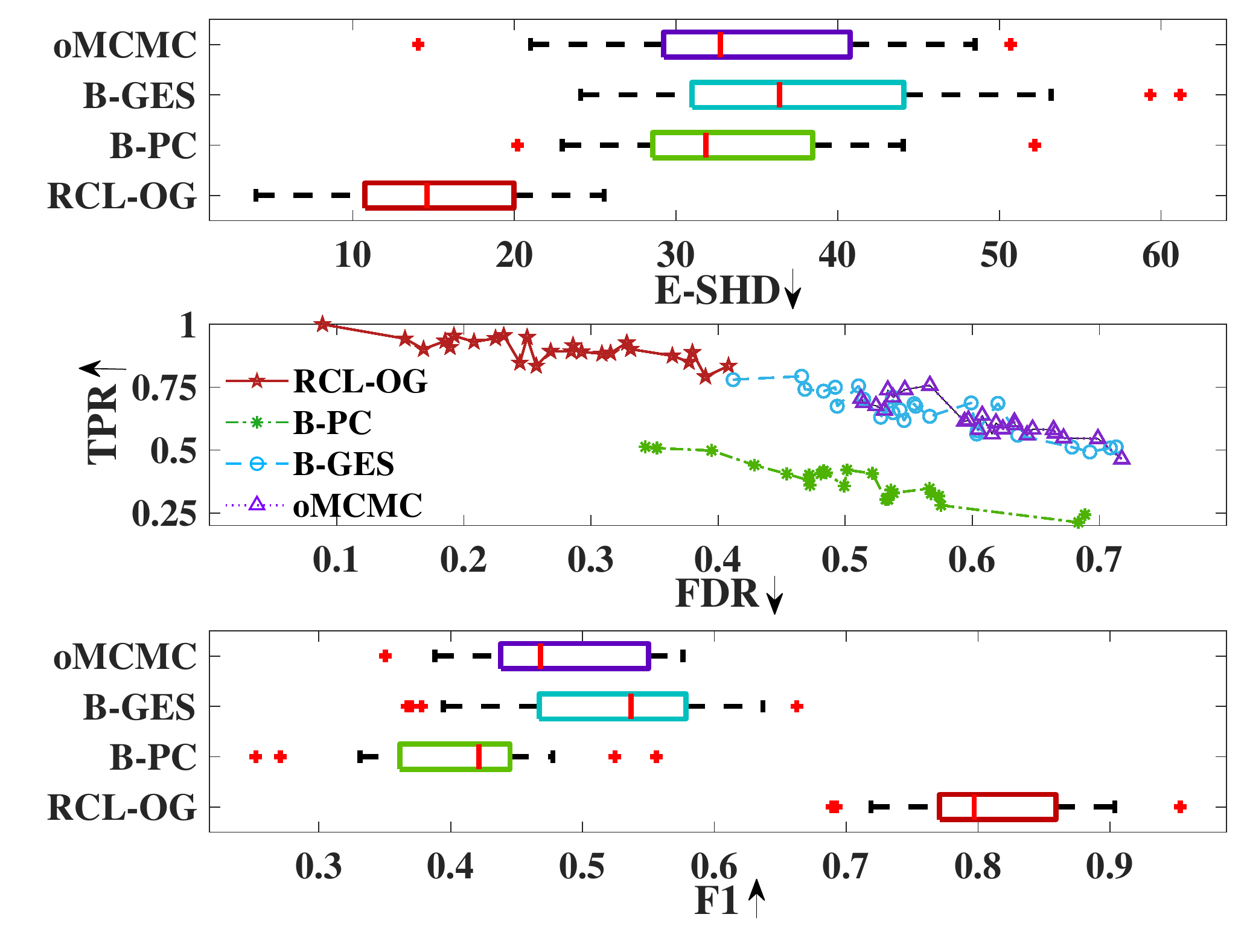} 
\caption{Results on DAGs with $d=20$ nodes. $\uparrow$/$\downarrow$ means the preferred direction of the metric value. 
}
\label{fig4}
\end{figure}

After Q-nets are trained, we can obtain the transition probability model and variable orderings by sampling on this model. We transform orderings into fully connected DAGs, get the weight of each edge on them by simple linear regression, and prune them by a threshold to get final DAGs. We perform experiments on graph with 20 nodes, and sample the ground-truth graphs following an Erd{\"o}s-R{\'e}nyi model, with $2d$ edges in expectation (a setting referred as ER2).  MCMC-based oMCMC \cite{friedman2003bayesian}, and traditional causal discovery methods, two variants of Bootstrapping \cite{lorch2021dibs} based on the score-based algorithm GES (B-GES, \cite{chickering2002optimal}), and the constraint-based algorithm PC (B-PC, \cite{spirtes2000causation}) are used as baselines.  We conduct experiments on 25 randomly generated different DAGs, and in each experiment, we generate a dataset of observations with 200 samples under the linear Gaussian model and sample 1000 DAGs from each method. We take the best 20 DAGs as the output of each method, and quantify the performance using the expected structural Hamming distance (E-SHD), which is approximately defined as the mean of SHD between the DAGs sampled by each method and the true graph:
\begin{equation}
    \displaystyle
    \mathrm{E-SHD}\approx\frac{1}{n}\sum^n_{k=1}\mathrm{SHD}(\mathcal{G}_k,\mathcal{G}^*)
    \label{eq13}
\end{equation}%
where $n$ is the number of sampled graphs $\mathcal{G}_k$, $\mathcal{G}^*$ is the true graph and $\mathrm{SHD}(\mathcal{G}_k,\mathcal{G}^*)$ is the structural Hamming distance between $\mathcal{G}_k$ and $\mathcal{G}^*$. We also report the line-plot of True Positive Rate (TPR) and False Discovery Rate (FDR) and the box-plot of F1-Score in Figure \ref{fig4}.

RCL-OG can more credibly sample the correct causal structure than other methods. The performance of B-PC and B-GES is restricted by their base algorithms PC and GES, so they cannot sample the exact graph structure. Compared with oMCMC, the state-action space of RCL-OG on order graph is much smaller. Therefore, with the trained transition model, RCL-OG is more likely to sample the correct order without high sampling cost. These experiments prove that RCL-OG can produce posterior probabilities close to those of topological orderings of ground-truth graph $\mathcal{G}^*$.

\begin{figure*}
\centering
\subfigure[Results of TPR]{
    \includegraphics[scale=0.37]{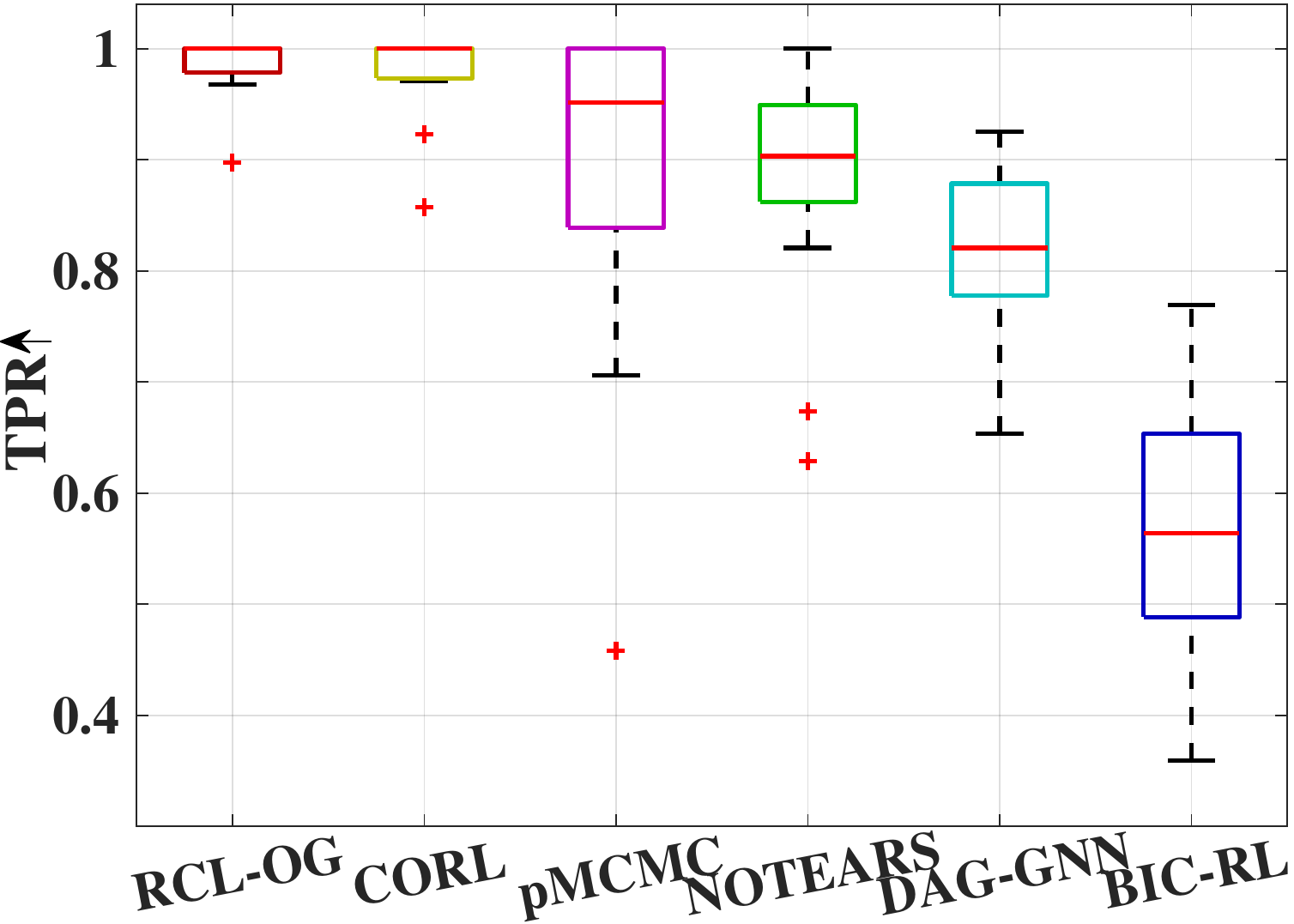}
}
\subfigure[Results of FDR]{
    \includegraphics[scale=0.37]{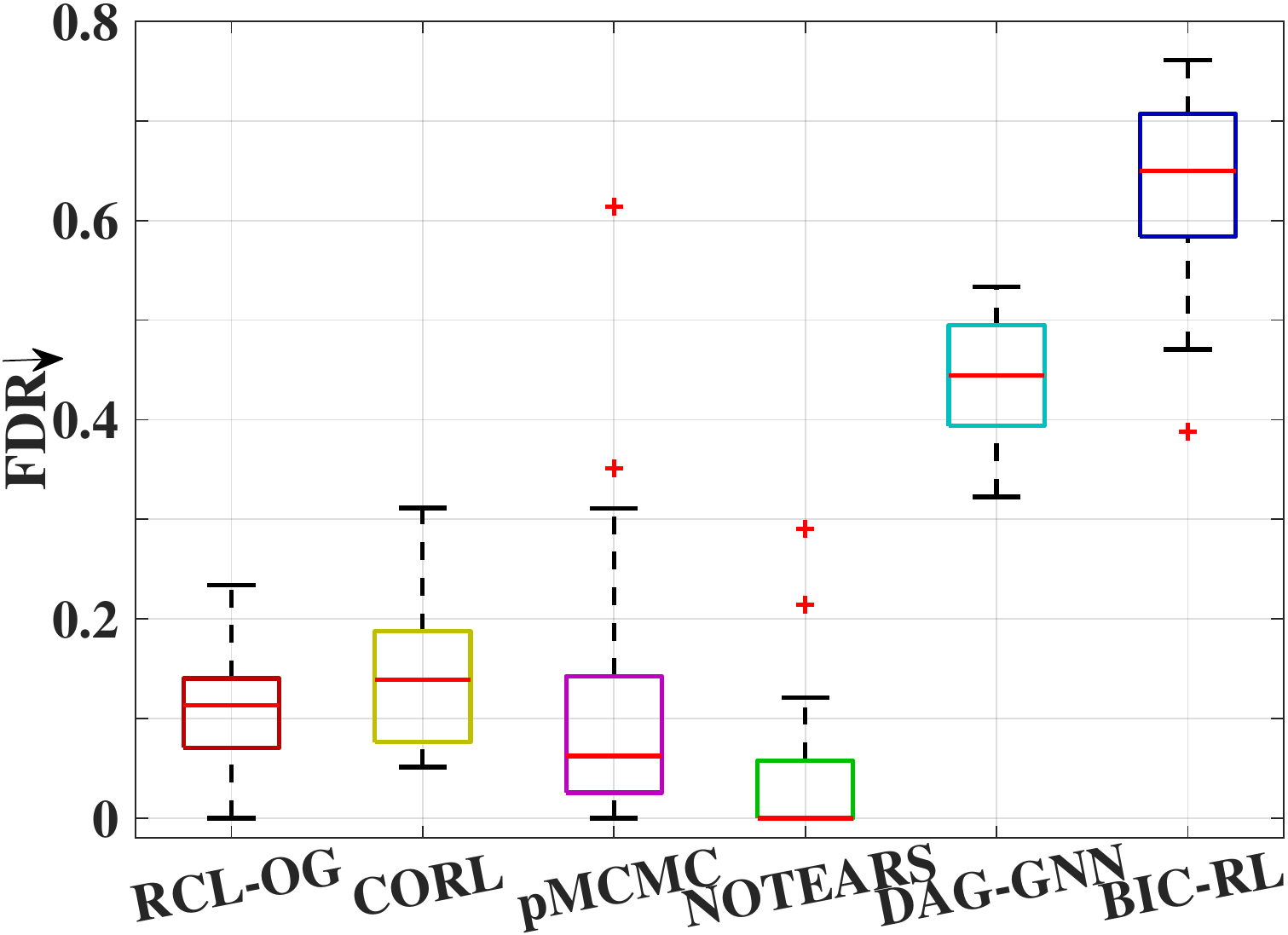}
}
\subfigure[Results of SHD]{
    \includegraphics[scale=0.37]{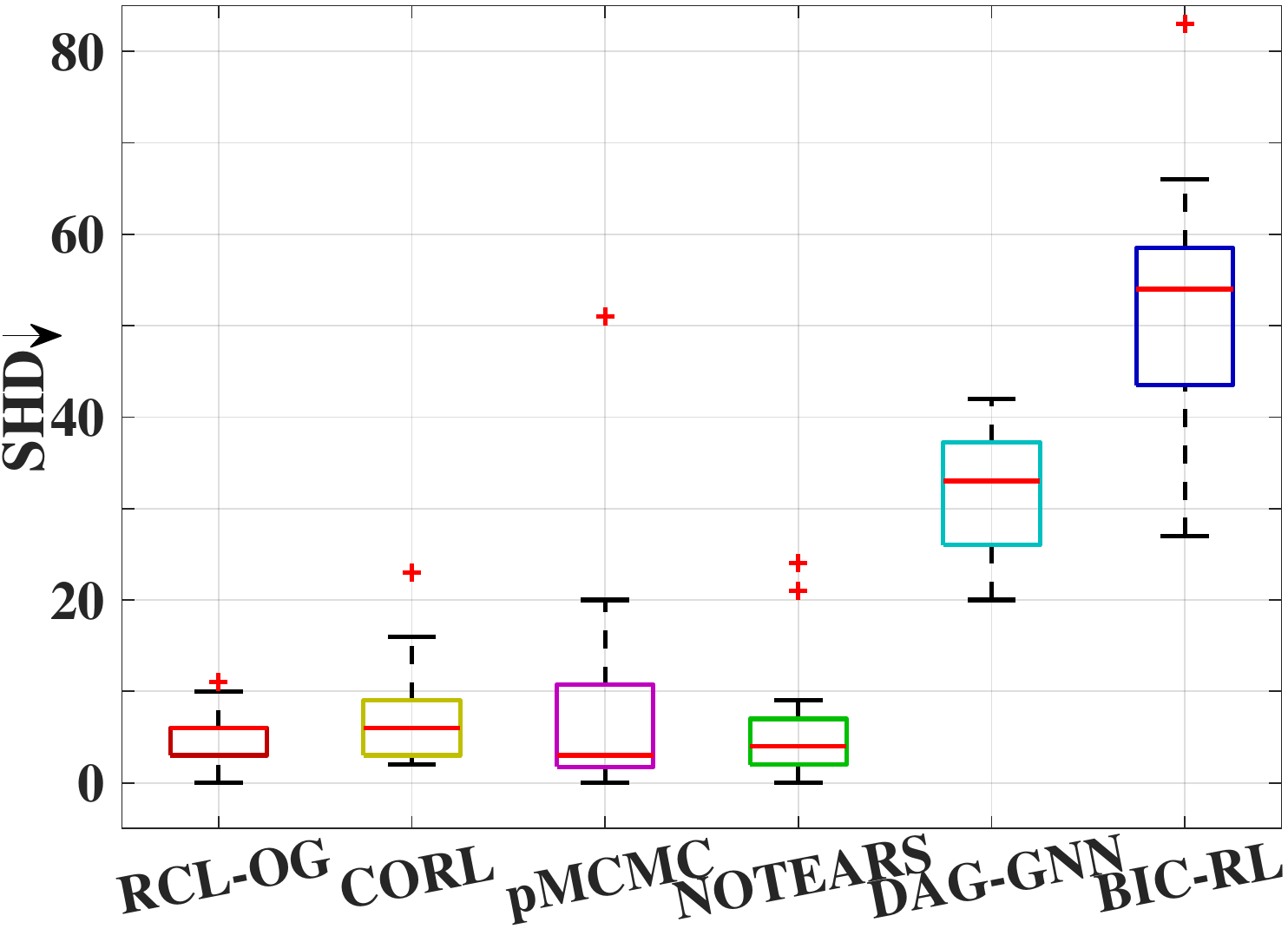}
}
\caption{Results on DAGs with $d=20$ nodes.}
\label{fig5}
\end{figure*}

Besides sampling possible orderings in the order graph space, RCL-OG can also get an exact single ordering. Starting from the initial state $s_0$ of the order graph, we can obtain a single ordering by selecting action with the maximum transition probability at each transition until the target state $s_d$. We process this ordering into a DAG in the same way as before and call it the final output of RCL-OG. We compare RCL-OG against with state-of-the-art causal discovery methods, each of which also outputs a single DAG. The baselines include NOTEARS \cite{zheng2018dags}, DAG-GNN \cite{yu2019dag}, BIC-RL \cite{zhu2019causal} and CORL \cite{wang2021ordering}. We also include an MCMC-based baseline pMCMC \cite{kuipers2017partition}. We use the same setup as the sampling experiment and report the TPR, FDR and SHD of each method in Figure \ref{fig5}. We have the following observations:\\
(i) RCL-OG clearly outperforms most compared methods across different evaluation criteria. BIC-RL has the worst performance in each metric, because it needs enough iterations to hit the correct causal structure in the whole DAGs space. DAG-GNN also performs poorly, because it uses graph neural network to reconstruct data, which requires a large amount of data for training. As a result, DAG-GNN maybe not able to learn a good causal structure with limited samples.  NOTEARS has a better performance than BIC-RL and DAG-GNN, since the adopted linear regression favors the loss function of NOTEARS.  pMCMC can also get the best results, but the nature of searching through probability-based sampling makes its results unstable. RCL-OG and CORL yield a similar performance, and achieve the best results on TPR, but slightly lose to NOTEARS on FDR. This is because they both prune the fully connected DAG induced from the ordering, but pruning often cannot remove all wrong edges. Even so, the results still prove that RCL-OG and CORL can find the correct topological ordering of causal graph. \\
(ii) RCL-OG has the highest TPR with ensured stability. By taking three metrics together, we can see that RCL-OG is slightly better than CORL in accuracy and stability. CORL searches the whole orderings space by RL and outputs the searched optimal ordering. In contrast, RCL-OG only needs to sample actions on the transition model with the maximum probability to get better (or similar) results. {The smaller search space and the deterministic sampling enables RCL-OG more stable than CORL.}\\
(iii) RCL-OG not only can find accurate causal topological orderings on a larger graph, but also has a higher search efficiency. We further compare RCL-OG against with CORL and pMCMC on larger graphs with $d$=40 nodes. With the
same experimental settings as before, we record the total number of searched orderings of each method in the training phase and in the sampling phase for attaining the similar TPR. We fix the maximum number of searches as $10^5$ to prevent them from indefinite search and record the counts in Table \ref{table1}. We find the results of pMCMC are less stable, and it occasionally fails to find an accurate DAG using the maximum number of searched orderings. We record its average number of searches when its TPR$\approx$0.97. For the case its TPR cannot reach $0.97$ within the maximum number of searches, we record its best TPR. We set TPR$\approx$0.97, because it is a common and good enough result for the three methods under the maximum number of searched. We see that both TPR and the number of used searches of pMCMC are worse than those of RCL-OG and CORL. The lower FPR of pMCMC is due to its detailed neighborhood selection, while RCL-OG and CORL simply use linear pruning. CORL can basically find a good enough ordering within the maximum number of searches, and it needs about $7.4$$\times$$10^4$ searches. To attach $\text{TPR}$$\approx$$0.97$, RCL-OG uses about $3.6$$\times$$10^4$ searched orderings, which is much smaller than CORL. That is because RCL-OG can explore a smaller state-action space on the order graph than CORL and pMCMC.

\begin{table}[ht]
\caption{Results on DAGs with $d=40$ nodes. \#search is the number of searched orderings for attaching the target TPR in the training and sampling phase.}
\centering
\begin{tabular}{c|r r r r}
\hline
                    & RCL-OG            & CORL              & pMCMC\\
\hline
TPR$\uparrow$           & 0.978             &  0.971            &  0.964\\
FPR$\downarrow$         & 0.049             &  0.055            &  0.016\\
\#search$\downarrow$    & 3.6$\times$$10^4$ &7.4$\times$$10^4$  &8.6$\times$$10^4$\\
\hline
\end{tabular}
\label{table1}
\end{table}

\subsection{Experiments on real-world data}

We evaluated RCL-OG on real-world flow cytometry data \cite{sachs2005causal} to learn a protein signaling causal network based on expression levels of proteins and phospholipids. This dataset is a benchmark for graphical model and widely used in the biological community. It contains continuous measurements of 11 phosphoproteins in individual T-cells. We select the observational data with 853 samples to infer causal structure, and take the DAG given by \cite{sachs2005causal} as the ground-truth graph, which contains 11 nodes and 17 edges.

We again use BIC as the score function, but due to the nonlinear nature of real data, we apply CAM pruning \cite{buhlmann2014cam} to obtain sparse graph structure. We adopt the same setup as the simulation experiments to evaluate the sampling performance of each algorithm and the capability of finding a single causal structure, respectively. We use more evaluation metrics on the found graph: number of total edges, number of correct edges, SHD and Structural Intervention Distance (SID) \cite{peters2015structural}. For the sampling performance, we report the average of these metrics. The experiment results are given in Table \ref{table2}.

\begin{table}[t]
\caption{Results of probability sampling (top) and causal structure learning (bottom) on the real dataset.}
\centering
\begin{tabular}{c|r r r r}
\hline
            & E-total   & E-correct$\uparrow$ & E-SHD$\downarrow$ & E-SID$\downarrow$\\
\hline
oMCMC       &7.00       &2.70       &16.85  &55.75\\
B-GES       &12.55      &3.95       &16.85  &59.05\\
B-PC        &11.95      &4.95       &13.10  &59.10\\
RCL-OG      &$\mathbf{10.30}$&$\mathbf{6.60}$&$\mathbf{11.70}$&$\mathbf{41.40}$\\
\hline
            & total     & correct$\uparrow$ & SHD$\downarrow$ & SID$\downarrow$\\
\hline
pMCMC       &7          &3          &14     &57\\
CORL        &10         &5          &13     &50\\
DAG-GNN     &12         &6          &13     &48\\
BIC-RL      &14         &7          &11     &58\\
NOTEARS     &15         &7          &11     &44\\
RCL-OG      &10 &7 &11 &\textbf{39}\\
\hline
\end{tabular}
\label{table2}
\end{table}

The sampling performance gaps between compared methods on real data are similar to those on simulated data, and RCL-OG shows obvious advantages over other sampling methods. As to the results of causal structure learning,  CORL and pMCMC has a poor performance, since they only search the high score structure, which may not be able to find the correct causal graph in real data. DAG-GNN gets a lower SID than CORL and pMCMC, but it outputs cycles. The results of NOTEARS and BIC-RL are similar, but NOTEARS has a lower SID, which indicates that NOTEARS finds a graph closer to the causal graph in terms of causal mechanisms. RCL-OG holds a better (or comparable) performance than any other compared method, and offers a good trade-off between more correct edges and total edges. In addition, the lowest SID indicates that RCL-OG restores a more accurate causal mechanism.

\section{Conclusions}
This paper studies how to accurately characterize the whole distribution over DAGs for effective causal structure learning using limited data.
We propose a new reinforcement learning based solution on variable order graphs with reduced spaces to effectively approximate the posterior distribution of orderings.
Experiments on simulated and real datasets prove that RCL-OG not only effectively approximates the posterior distribution of ground truths, but also stably and accurately finds the single causal structure.

\end{document}